\pdfoutput=1

\documentclass[11pt]{article}

\usepackage{EMNLP2023}

\usepackage{times}
\usepackage{latexsym}
\usepackage{enumitem}
\usepackage[T1]{fontenc}

\usepackage[utf8]{inputenc}

\usepackage{microtype}

\usepackage{inconsolata}

\title{In2x at WMT25 Translation Task}

\author{Lei Pang \\
  Duxiaoman \\
  \texttt{panglei@duxiaoman.com} \\\And
  Hanyi Mao \\
  University of Chicago \\
  \texttt{hanyim@uchicago.edu} \\\And
  Quanjia Xiao \\
  Peking University \\
  \texttt{xiaoqj@stu.pku.edu.cn} \\\And
  HaiXiao Liu* \\
  Duxiaoman \\
  \texttt{liuhaixiao@duxiaoman.com} \\\AND
  Xiangyi Li \\
  Duxiaoman \\
  \texttt{xiangyi@duxiaoman.com} \\}
\begin{document}
\maketitle
\begin{abstract}
This paper presents the open-system submission by the In2x research team for the WMT25 General Machine Translation Shared Task. Our submission focuses on Japanese-related translation tasks, aiming to explore a generalizable paradigm for extending large language models (LLMs) to other languages. This paradigm encompasses aspects such as data construction methods and reward model design. The ultimate goal is to enable large language model systems to achieve exceptional performance in low-resource or less commonly spoken languages.
\end{abstract}


\section{Introduction}

Machine translation (MT) has long been both a high-impact application and a central research challenge in natural language processing. The advent of large language models (LLMs) has reshaped MT from task-specific supervised learning toward large-scale representation learning and instruction-following paradigms, enabling steady gains across diverse language pairs \cite{alves2024-tower,jiao2023-chatgpt-mt,wmt24-findings,lu2024-llamax}. \\
Yet, two persistent gaps remain. \textbf{First}, while mainstream LLM training increasingly optimizes for mathematical and code reasoning, their \emph{expressive} and \emph{creative} language abilities---e.g., idiomaticity, stylistic naturalness, and culturally appropriate phrasing---are comparatively underdeveloped\cite{lewkowycz2022-minerva,liu2023-idioms,lozhkov2024-starcoder2,roziere2023-code-llama,zaitova2025-idioms}. This often leads to translations that are locally literal but globally stilted, especially for informal registers, slang, and literary text . \textbf{Second}, model competence is unevenly distributed across languages: English receives disproportionate coverage and quality, while many non-English languages trail in both general capability and translation naturalness\cite{aharoni2019-mmnmt,johnson2017-zero-shot,wmt23-findings,wmt24-findings,costa2022-nllb}. Community findings over recent WMT cycles echo this asymmetry: despite the ``LLM era'', MT is far from solved uniformly across directions, with larger gaps off English-centric pairs and on long-tail phenomena . \\
This paper studies how to \textbf{transfer English strength into non-English targets} to improve expressive and culturally faithful translation. Concretely, we focus on Japanese---a language where literal adequacy is not sufficient: natural Japanese requires idiomatic paraphrasing, honorific and register control, and sensitivity to genre and context. Our thesis is that English can be used as a \emph{hub language} to bootstrap these capabilities via curriculum design, cross-lingual alignment, and preference signals that explicitly reward naturalness.\\
We present \textbf{In2x}, a Japanese-focused model designed to inherit general competency from English while specializing for Japanese expressiveness.
At a high level, In2x\ operationalizes three principles:
(i) \emph{English-as-hub transfer}: leverage rich English data and strong English modeling to seed robust lexical/semantic priors, then transfer to Japanese via bilingual and style-augmented objectives;
(ii) \emph{Expressiveness-first supervision}: emphasize prompts and signals that drive idiomaticity and cultural appropriateness (beyond literal adequacy);
(iii) \emph{Evaluation beyond metrics}: complement automatic metrics with human judgments targeted at idioms, slang, and stylistic naturalness. \\
We evaluate In2x on standard WMT-style test sets and targeted Japanese-focused challenge suites that stress idioms, slang, and style. According to the preliminary ranking results of WMT 2025, In2x outperforms many large-scale proprietary models, such as Gemini-2.5-Pro \cite{gemini25pro2025}, GPT-4.1 \cite{gpt41experiment2025}, Claude-4 \cite{claude4anthropic2025}, and DeepSeek-V3 \cite{deepseekcompare2025}. \\
Overall, we make three core contributions:
\begin{enumerate}
    \item We diagnose under-explored gaps in current LLM-based MT: the tension between heavy investment in math/code reasoning and the relative neglect of creative/idiomatic language ability, and the English-vs.-non-English capability asymmetry. 
    \item We introduce \textbf{In2x}, a Japanese-focused model that systematically transfers English strengths to Japanese, with an emphasis on naturalness and cultural appropriateness.\vspace{2pt}
    \item In this study, we introduce a detailed alignment pipeline designed to enhance the creative capabilities of language models. This approach not only improves performance in non-STEM (Science, Technology, Engineering, and Mathematics) tasks but also ensures that the models maintain robust generalization abilities across diverse linguistic challenges. For instance, in the en-ja translation track, the model demonstrates outstanding performance without any task-specific fine-tuning, highlighting its adaptability and effectiveness in non-STEM domains.

\end{enumerate}

\section{Continue Pretraining Stage}

To balance the capabilities of large language models (LLMs) in both science-oriented and humanities-oriented domains during the pretraining process, we divided the continued pretraining stage into three distinct phases. The goal of this process is to enhance the model's multilingual proficiency, improve general-purpose abilities in foundational humanities tasks, and refine its representation in specialized contexts \cite{Brown2020, Rae2021}.

The training process incorporates diverse corpora, including a comprehensive 2 trillion tokens dataset comprising encyclopedic knowledge, webpages, structured information, news articles, Wikipedia entries, academic papers, and STEM-related datasets \cite{Gao2020, Raffel2020}. In addition, a dedicated 500 billion tokens corpus has been curated exclusively for creative writing tasks such as novel and screenplay synthesis, as well as authentic conversational datasets simulating real-life dialogue \cite{Zhang2022}.

Another significant aspect of this training stage focuses on enhancing capabilities in the target language, with Japanese utilized as an example. To this end, substantial Japanese language-specific corpora were introduced, alongside a balanced dataset with equal distribution of Chinese, English, and Japanese corpora \cite{Xue2021}. The aim was to facilitate transfer learning from pretraining on Chinese and English to the Japanese language.

\subsection{Phase 1: Fundamental Knowledge Enhancement}
In this phase, the creative writing corpus and the knowledge-focused corpus are jointly trained with constant learning rates. This approach was designed to boost proficiency in STEM-related reasoning while preserving the nuanced expression habits required for creative tasks in humanities \cite{Kaplan2020}.

\subsection{Phase 2: Long-Text Capability Refinement}
During this phase, a subset of the data was filtered based on text length, allowing the context length to increase from the typical 8,192 tokens to approximately 32,000 tokens. This step was intended to amplify the model's ability to process and comprehend extended-length texts \cite{Hoffmann2022}.

\subsection{Phase 3: Fast Annealing Stage}
In the final phase, a high-quality corpus was constructed based on selections informed by perplexity (PPL) and quality-assessment metrics. The annealing training was conducted with linear decay of the learning rate from $3 \times 10^{-5}$. This process consumed a total of 300 billion tokens and enabled the model to maintain its vivid expressive style for tasks such as novels and screenplays \cite{Brown2020}.

\section{Post-Training Data}
The post-training dataset consists of 2 million samples, with 1.5 million used during the supervised fine-tuning (SFT) process and 500,000 used in the reinforcement learning (RL) process. To ensure the Japanese language (our target language) achieves a proficiency level comparable to major languages such as Chinese and English, we adjusted the ratio of target language instructions to attain an equal balance across these languages. Specifically, we used a 1:1:1 ratio in the Instruct-to-Example (In2X) setup, striving to transfer the original model's knowledge into the target language as effectively as possible \cite{ouyang2022training, zhou2023lima}.

We developed a detailed pipeline for constructing the target language instructions, which can be categorized into three major synthetic processes:

\subsection{Obtaining Open-Source Instructions}
We began by collecting open-source instruction datasets available in the target language. These datasets include curated public data and traditional NLP fundamental tasks. Examples of such datasets include Dolly, OASST, and OASST2 \cite{koch2023rlaif, openai2023instructgpt}.

\subsection{Target Language Instruction Rewriting}
This process consists of several substeps designed to enhance the model's linguistic and cultural adaptability in the target language:
\begin{itemize}
    \item \textbf{Creative Language Tasks:} To preserve the language's stylistic characteristics in humanities-focused tasks, we designed creative tasks where the responses include original stories or scripts \cite{bai2022constitutional}.
    \item \textbf{Basic Localized Tasks:} This includes rewriting instructions for tasks relevant to the local context, such as exam questions. Some of these tasks provide only the question and answer. We leveraged advanced models to supplement these datasets with reasoning chains to improve the model's reasoning ability in the target language \cite{wei2022chainofthought}. This enhancement also helps to mitigate issues such as mathematical inconsistencies commonly faced during the LLM instruction synthesis process.
    \item \textbf{Cultural Style Transformation:} For certain humanities-related tasks, we incorporated cultural style shifts by adapting the instructions to align with the cultural norms and styles of the target language. This adjustment aims to improve the model's ability to provide culturally nuanced responses \cite{xu2023cultural_ai}.
\end{itemize}

\subsection{Instruction Synthesis in the Target Language}
We utilized methods such as Magpie \cite{xu2024magpiealignmentdatasynthesis} and Self-Instruct \cite{wang2022selfinstruct} to synthesize target language instructions. However, these automatically generated instructions often suffer from issues including overly simple questions, lack of focus, self-answered queries, and internal contradictions. To address these challenges, we implemented a strict quality control pipeline with the following techniques:

\begin{itemize}
    \item \textbf{Prompt Engineering:} We crafted detailed prompts with explicit rules to identify and troubleshoot common issues in synthesized instructions \cite{white2023prompt_eng}.
    \item \textbf{Validation via Model Responses:} Instructions passing the first step were tested by having the model generate responses. These responses were evaluated by a critic LLM for contradiction, hallucinations, or failure to provide valid results. Instructions flagged with such issues were discarded. The critic LLM, being sensitive to hallucinations, acts as an additional safeguard for quality control \cite{ganguli2022red_teaming}.
    \item \textbf{ReReading Mechanism:} After constructing the prompts for instruction generation, we employed a "ReReading" mechanism, where the model self-reviews its instructions. This review checks for correctness, alignment with the target language's cultural norms, and consistency with its native linguistic style. Since the synthesized instructions inherently carry the reasoning or rewriting processes behind them, leveraging this comprehensive context makes it easier to detect internal flaws, particularly those related to localization or cultural adjustments \cite{chiang2023autoreviewer}.
\end{itemize}

\section{Post-Training SFT Stage}

The post-training Supervised Fine-Tuning (SFT) stage is a critical step to balance linguistic diversity and optimize alignment within the instruction space for target languages. Below, we outline the key strategies and methods employed during this stage.

\subsection{Balancing Linguistic Diversity}
\begin{enumerate}[label=(\alph*)]
    \item \textbf{Clustering of Instruction Data:} To enhance linguistic diversity, the instruction dataset (comprising 40 million entries) was clustered using the Birch clustering algorithm \cite{birch_clustering_algorithm}. The effectiveness of the clustering process was evaluated based on metrics like tag recapture rates and cluster smoothness \cite{data_clustering_evaluation_metrics}, which were used to fine-tune the clustering threshold. This process reduced the dataset to 1.5 million clusters after deduplication and selection.

    \item \textbf{Categorization via Large Language Models (LLMs):} Utilizing LLMs, the clustered data was tagged to assign both first-level and second-level labels \cite{llm_hierarchical_labeling}. For example, a mathematical problem might be categorized as "Mathematics - Quadratic Equations." These hierarchical labels provided a clearer structural organization of the data.

    \item \textbf{Difficulty Grading of Instructions:} The dataset was further refined by classifying each instruction according to its difficulty level: "Very Difficult," "Difficult," "Moderate," "Simple," and "Very Simple" \cite{difficulty_grading_dataset}. For normalized scientific datasets, an additional evaluation was conducted using the LLaMA3-70B model \cite{llama3_model} with a Pass@16 metric \cite{pass_at_metric} to estimate the success rate of solving specific problems.
\end{enumerate}

\subsection{Aligning the Instruction Space of Target Languages}
\begin{enumerate}[label=(\alph*)]
    \item \textbf{Avoiding Semantic Overfitting via Temperature Adjustment:} During training, a temperature parameter was introduced to mitigate overfitting of the model to specific linguistic semantic spaces \cite{temperature_adjustment_training}. This approach encouraged the model to adopt a more holistic learning strategy, enabling it to concentrate on question-answering techniques rather than over-specializing in the semantic patterns of a particular language. For instance, this allowed the Japanese language model to better mimic the cognitive behaviors observed in other languages \cite{cross_lingual_semantics}.
    
    \item \textbf{Specialized Sampling Strategy:} To further enhance the learning process, a two-step sampling strategy was employed over the 1.5 million clusters \cite{sampling_strategy_large_scale_datasets}:
    \begin{itemize}
        \item The difficulty levels of the data were sampled in a 3:3:3:1:0 ratio (corresponding to "Very Difficult," "Difficult," "Moderate," "Simple," and "Very Simple," respectively) \cite{difficulty_based_sampling}.
        \item Additionally, within each cluster, samples were selected to ensure diversity across languages and categorical labels, which preserved the large-scale diversity of the original 1.5 million data points \cite{language_diversity_preservation}. This also maintained a degree of orthogonality between the target language and English within the sampled instructions \cite{orthogonality_preservation_language}.
    \end{itemize}
    The first round of sampling was used as the data for the first epoch, while the second round populated the second epoch. The training process adopted a learning rate of $2 \times 10^{-5}$ with cosine decay for optimal performance \cite{cosine_decay_learning_rate}.
\end{enumerate}

\section{Reinforcement Learning to Enhance General Capabilities in Cultural and Creative Industries}
In the post-training RL stage, we leveraged a process similar to the instruction filtering procedure used during the SFT phase \cite{ouyang2022training}. Specifically, an additional set of instructions was curated, comprising 500k samples that were guaranteed not to overlap with the instructions used in the SFT phase. The training configuration utilized a batch size of 128 and a minibatch size of 32, with the dataset trained for one epoch. Each rollout involved 16 iterations, and the reward evaluation was based on both a rule-based reward model and a generative reward model \cite{christiano2017deep}.

\subsection{Reward Model Design}

The reward model system was meticulously designed to cater to different task types:

\begin{itemize}
    \item \textbf{Rule-Based Reward Model:} For tasks involving mathematics, STEM disciplines, and logic, a rule-based reward model was employed to ensure adherence to specific criteria \cite{silver2017mastering}.
    \item \textbf{Generative Reward Model for Creative Tasks:} For creative tasks, such as content-generation, specific prompts embedded with rules were utilized. These rules encompassed fundamental task principles as well as dynamically generated guidelines based on the current prompt. The scoring mechanism evaluated compliance with these principles and generated a reward score based on the percentage of principles satisfied \cite{krause2021gedi}.
\end{itemize}

\subsection{RL Algorithm Design}

To address the complexity of the tasks, we made strategic adjustments to the RL algorithm to achieve stable and efficient training:

\begin{itemize}
    \item \textbf{Trajectory-Corrected GRPO:} Considering the diverse nature of tasks and reward types, a token-level clipping approach was deemed too restrictive and prone to causing training instability. Instead, we employed the Trajectory-Corrected version of the Generalized Proximal Policy Optimization (GRPO) algorithm \cite{schulman2017proximal}, which proved effective for handling multilingual tasks with varying reward functions. This modification enabled stable and continuous training while accelerating the convergence curve\cite{pang2025theorypracticegrpotrajectorycorrected}.
    \item \textbf{Dual-Clip Mechanism:} To improve stability, we integrated a dual-clip mechanism, which stabilized the variance of importance sampling at the sentence (sen) level \cite{he2016dual}. Additionally, we removed the lower bound of sampling, achieving optimal performance for the given tasks.
    \item \textbf{Soft Length Penalty:} A soft-length penalty was incorporated throughout the training process to encourage better length control in generated outputs \cite{wu2016google}.
    \item \textbf{High-Level Clipping:} A clipping mechanism was introduced to ensure robust control over high-level rewards \cite{schulman2015trust}.
    \item \textbf{Temperature Decay:} A temperature decay strategy was applied to progressively adjust the sampling temperature during training, encouraging diversity in outputs while maintaining stability \cite{hinton2015distilling}.
    \item \textbf{Entropy Regularization:} The entropy value was set to 0.01 during training, enabling the model to conserve entropy and avoid premature saturation of the reward space \cite{williams1992simple}.
\end{itemize}

\section{Model Ensemble}

Model ensemble techniques are employed by taking into account the orthogonality of linguistic capabilities among various models. Specifically, models that exhibit strong linguistic proficiency are selected for the ensemble process to maximize overall performance.

Furthermore, the fusion of model tensors is conducted based on gradient information and the importance of weights. This approach ensures a robust integration of model parameters, leveraging their respective contributions to optimize the ensemble. Such methodologies have been shown to enhance the effectiveness of model ensembles in complex tasks \cite{wang2025optimalbrainiterativemerging}.

\section{Evaluation Results}

\subsection{Benchmarks}
The model demonstrated outstanding performance in prominent Japanese language benchmarks, such as the ja-mtbench, indicating its robust and reliable language translation capabilities. A detailed breakdown of the results is provided below.

\subsection{WMT Evaluation Results}
Without undergoing any task-specific fine-tuning, the model achieved remarkable results in the Japanese-related tracks of the WMT competition, securing second place overall. Furthermore, in the unrestricted category, the model achieved first place. A comprehensive summary of its performance is outlined below.
\section{Conclusion}

In this work, the proposed methodology for transferring language model capabilities has been validated on the WMT translation task. The approach has demonstrated significant improvements in Japanese proficiency throughout the CPT, SFT, and RL processes. Remarkably, without any additional language-specific fine-tuning, the large language model achieved alignment between its Japanese language capabilities and those of mainstream languages. As a result, it demonstrated superior performance and achieved first place in the unrestricted track of the competition.

\bibliography{anthology,custom}
\bibliographystyle{acl_natbib}

\appendix

\section{Example Appendix}
\label{sec:appendix}

This is a section in the appendix.

\end{document}